\RequirePackage{fix-cm}
\documentclass[smallextended]{svjour3}       
\smartqed  
\usepackage{graphicx, booktabs, natbib}
\usepackage{float}
\usepackage{url}
\usepackage{multirow}
\begin{document}

\title{Lahjoita puhetta -- a large-scale corpus of spoken Finnish with some benchmarks
}
%


\titlerunning{Lahjoita puhetta}        

\author{ Anssi Moisio*$^1$ \and Dejan Porjazovski$^1$ \and Aku Rouhe$^1$ \and Yaroslav Getman$^1$ \and Anja Virkkunen$^1$ \and
        Tam\'as Gr\'osz$^1$ \and
        Krister Lindén$^2$ \and
        Mikko Kurimo$^1$ 
}

\authorrunning{Moisio et al.} 

\institute{*Corresponding author \\
    \at $^1$Department of Signal Processing and Acoustics, Aalto University, Espoo, Finland \\
              $^1$\email{firstname.lastname@aalto.fi}           
      \at $^2$Department of Digital Humanities, University of Helsinki, Finland \\
      $^2$\email{firstname.lastname@helsinki.fi}
}

\date{Received: date / Accepted: date}

\maketitle

\begin{abstract}
The Donate Speech campaign has so far succeeded in gathering approximately 3600 hours of ordinary, colloquial Finnish speech into the \textit{Lahjoita puhetta} (\textit{Donate Speech}) corpus. The corpus includes over twenty thousand speakers from all the regions of Finland and from all age brackets. The primary goals of the collection were to create a representative, large-scale resource to study spontaneous spoken Finnish and to accelerate the development of language technology and speech-based services. In this paper, we present the collection process and the collected corpus, and showcase its versatility through multiple use cases.
The evaluated use cases include: automatic speech recognition of spontaneous speech, detection of age, gender, dialect and topic and metadata analysis. We provide benchmarks for the use cases, as well downloadable, trained baseline systems with open-source code for reproducibility.
One further use case is to verify the metadata and transcripts given in this corpus itself, and to suggest artificial metadata and transcripts for the part of the corpus where it is missing.
\end{abstract}

\section{Introduction}
\label{intro}
The preservation of spoken colloquial language is an important task, 
which requires the collection of relevant materials and their careful curation. The Donate Speech (Lahjoita puhetta) campaign embarked on the quest of preserving the current state of the spoken Finnish language and boosting the development of AI that understands spoken Finnish. To this end, a large collection campaign was initiated that resulted in the creation of a large-scale colloquial Finnish speech corpus. In this paper, we explain how the collection and curation of the data were performed to maximise the amount participants while still ensuring a high quality of the dataset. Furthermore, we will also demonstrate with pilot projects and their results how the materials can be used to study and develop new technology and services in the Finnish language.

Currently, there is only one large freely available transcribed Finnish speech corpus, the Finnish Parliament ASR Corpus\footnote{\url{urn.fi/urn:nbn:fi:lb-2021051903}}.
It contains over 3000 hours of professionally transcribed speech which is rather formal in style and often read from the speaker's notes.
However, colloquial, spontaneous Finnish differs significantly from formal Finnish in multiple aspects. Considering phonological features, for instance, durations of phones are longer in read speech than in spontaneous speech \citep{lennes2009segmental}. From the morphological and lexical point of view, it is common to truncate or combine words, and to use incorrect word inflections in addition to words not used in written text. Since Finnish has a near-phonemic orthography, the differences can be transcribed mostly unambiguously into text, and there is no fixed correct transcription style. Therefore, the phonological spelling variations of a single word can be numerous, since different pronunciations of a word can easily be rendered in written form, which further increases the distance between the domains of formal and colloquial Finnish.

There are a few smaller corpora that include carefully transcribed spontaneous, colloquial Finnish speech. The SPEECON \citep{iskra2002speecon} corpus is a collection of speech for multiple languages, recorded in varying environments. It includes both read and spontaneous speech from 550 speakers. The spontaneous Finnish part includes 10 sentences from each speaker, in total about 18.8 hours. The FinDialogue\footnote{\url{urn.fi/urn:nbn:fi:lb-2016041421}} part of the FinINTAS \citep{lennes2009segmental} corpus contains 6338 utterances by 22 speakers. The speech is from spontaneous and unmonitored conversations between participants, and includes about 10.4 hours of speech in total. The DSPCON\footnote{\url{urn.fi/urn:nbn:fi:lb-201708251}} corpus consists of free-form conversations between students, recorded at the Aalto University between 2013 and 2016. It includes 5281 spontaneous sentences from 218 different male students and 24 female students, totalling 9.8 hours \citep{enarvi2018modeling}. Combining these three corpora, there are about 40 hours of transcribed spontaneous Finnish speech currently 
available for research (non-commercial) use\footnote{While SPEECON is quite expensive, the other two corpora are free.}, to the best of our knowledge. 
We note that substantial amounts of Finnish colloquial speech has been collected in the 1960s and 1970s by the National Institute for the languages of Finland as well as some cultural foundations, but that data is not yet available for commercial development use according to the European data protection legislation.

For major languages like English, large spontaneous and colloquial speech corpora are available for research and commercial use.
The Switchboard corpus \citep{godfrey1992switchboard} consists of about 260 hours of telephone conversations among 302 male and 241 female speakers. The Fisher corpus \citep{cieri2004fisher} includes approximately 2000 hours of colloquial telephone conversations. These two corpora, for example, have been actively used in speech research for many years now, and technologies built for spontaneous English have greatly benefited from the datasets. 
Even though Finnish has far fewer speakers than the major languages (not even in the top 100), the new \textit{Lahjoita puhetta} corpus covers many more speakers per language than probably any other publicly available spontaneous speech corpus. 

All of the tools and resources described in this work can be accessed online\footnote{\url{github.com/aalto-speech/lahjoita-puhetta-resources}}. The contributions of this work include:
\begin{enumerate}
    \item An open large colloquial speech data set for Finnish
    \item A successful concept for large-scale speech data curation
    \item Relevant benchmarks for speech, gender, age, dialect and topic recognition
    \item Trained, downloadable baseline systems for the benchmarks, and open source code for reproducing the systems
\end{enumerate}

\section{Data Collection}


The \textit{Lahjoita puhetta 2021} release consists of 3600 hours of data out of which about 1600 hours have been transcribed. The data covers all regions of Finland and has both male and female, mostly native, speakers in all age brackets.

The speech material donated during the campaign is shared by the Language Bank of Finland (Kielipankki)\footnote{\url{kielipankki.fi}}, coordinated by the University of Helsinki. Since speech samples may contain personal data, they are protected by European and national data protection legislation, most notably by the General Data Protection Regulation (GDPR)\footnote{Regulation (EU) 2016/679 of the European Parliament and of the Council of 27 April 2016}. The speech material has been collected based on the legitimate interest of individual researchers, universities, research organisations and private companies to study language or artificial intelligence, to develop AI solutions and to provide higher education in the aforementioned areas.

To inform the individuals who donated their speech to the campaign, two essential documents were drafted: a short information page including simple conditions of participation, and a more comprehensive data protection policy. To use legitimate interest as the lawful basis of the processing of personal data, it was necessary to accomplish a balance test to ensure that the legitimate interests are not overridden by the interests or fundamental rights and freedoms of the data subject. It was considered that the risks to the rights and freedoms of natural persons are rather low, but to be sure, a data protection impact assessment (DPIA) was also made. For a more detailed descriptions of the campaign and its legal background, see \citep{lindenCLARINbook}.

The goal of the campaign was not merely to collect a vast amount of any kind of speech, but to reach out to as many different groups of Finnish speakers and to as many individuals as possible. In marketing the campaign to citizens, it was emphasised that all variants of spoken Finnish are welcome, including speech from second language Finnish learners. However, in order to understand the privacy notice and the instructions, a certain level of language proficiency was required from the speech donors. 

Key issues and challenges for the design of the user interface were in determining elicitation methods that entice a person to speak freely, gaining the trust of the speaker, making him feel comfortable while also satisfying legal constraints for presenting enough required information in an easy to understand format, as well as more technical choices of supported platforms, presentation forms, visual and auditory feedback of the on-going recording or its quality. After some ideas for themes had been formulated and tested, Yle (the Finnish Broadcasting Company) settled on the fail-safe recurring functions of showing a video, a picture or some textual content enticing a person to speak with an easy-to-use one-button starting and stopping of the recording. 

\begin{figure}[htb]
    \centering
    \includegraphics[width=0.9\linewidth]{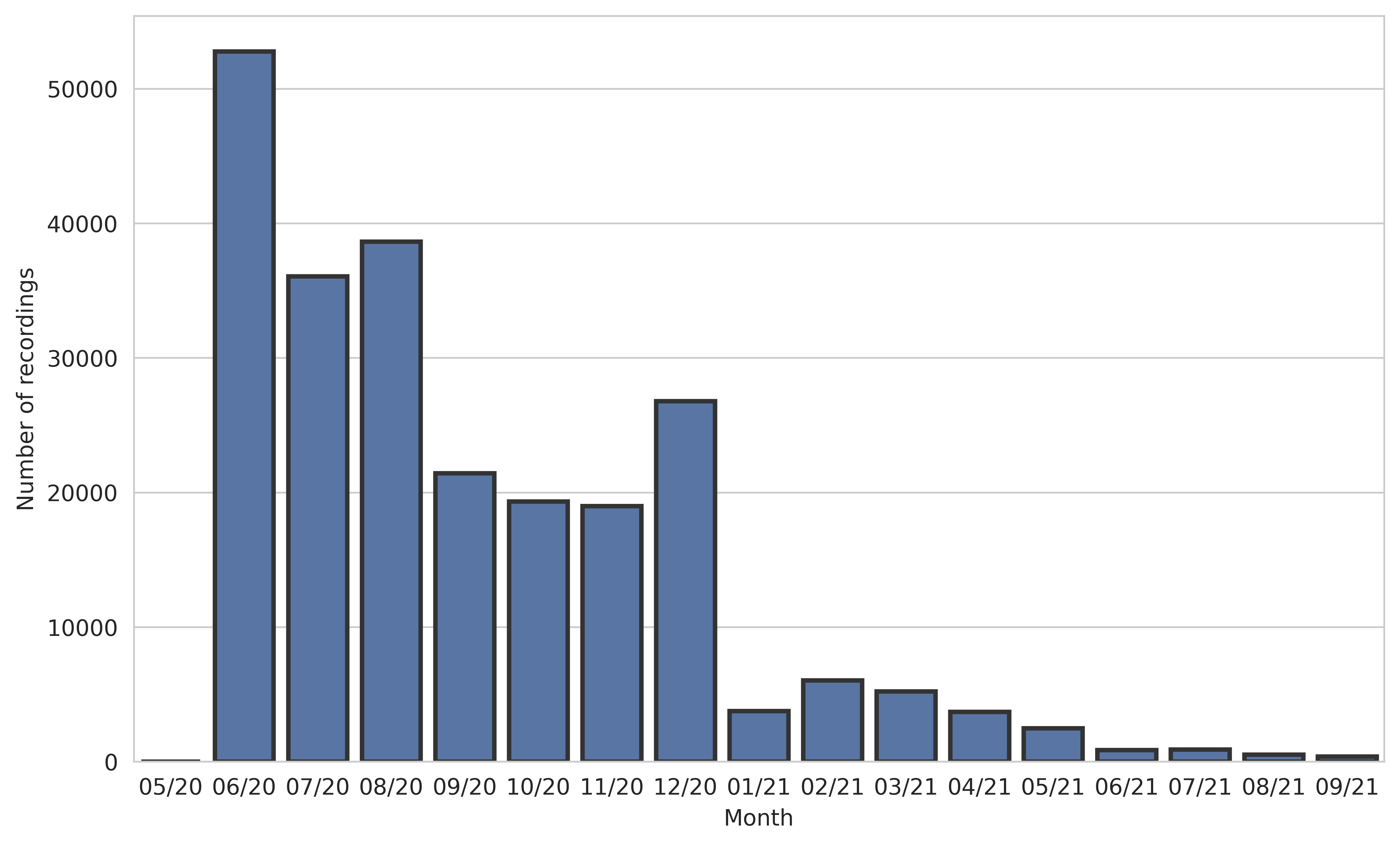}
    \caption{The number of recordings received in each month during the campaign.}
    \label{fig:recs_month}
\end{figure}

Cooperating with Yle was crucial for the marketing of the campaign and for attracting the attention of the Citizens of Finland for the campaign. In the end, Yle developed around 40 straightforward topics, within ten different themes, for stimulating the collecting of speech data. As part of the campaign, Yle made comical infomercials with requests to the general public to donate speech. These were broadcast during programme breaks in national radio and TV channels during the summer and autumn of the Covid-19 pandemic in 2020 with some trailing reruns during spring 2021.
In 2021 the data collection campaign was awarded the best European Digital Audio Project prize by PRIX EUROPA, which was founded by the European Parliament, the European Commission and the European Cultural Foundation in 1987. 

To illustrate the campaign results with regard to collection speed, the number of recordings received each month during the campaign is shown in Figure \ref{fig:recs_month}. The peaks in the beginning and at the end of 2020 reflect the effects of the increased public advertising activity. 

\subsection{Metadata complementing the speech corpus} \label{sec:metadata}
Identities of speakers were not collected explicitly, but we assume that one application client identity (of the browser or smart phone application used for recording) corresponds to one speaker. This assumption is not watertight since one person may use multiple application clients, or multiple persons may use one client, but the correspondence generally holds.
Assuming this, the number of speakers is well over 20k, which means quite a good sample of Finnish speakers, which are fewer than six million in total.  

Opening the Lahjoita puhetta website or phone app, the user is offered a few different themes to choose from. To focus the campaign, all of the themes are not always available on the website. The complete list of themes, and their English translations and abbreviations used in this text, is the following:
\begin{itemize} 
    \item “Eläinystävät” (“Animal friends”, A)
    \item “Urheiluhetket” (“Sports moments”, SP)
    \item “K-18” (“Rated R”, R)
    \item “Luonto, sää ja mää” (“Nature”, N)
    \item “Lähelläni juuri nyt” (“My surroundings”, M)
    \item “Mediataidot 4-6 lk.” (“Media skills – grade 4-6”, MS4)
    \item “Mediataidot 8-9 lk.” (“Media skills – grade 8-9”, MS8)
    \item “Mediataidot lukio” (“Media skills – high school”, MSH)
    \item “Kirottu korona” (“The cursed covid”, C)
    \item “Sukella kesään” (“Summer”, S)
\end{itemize}
Each theme includes up to eight different topics that ask a question or in some other way invites the user to speak about the topic. Each recording therefore pertains to some general theme, as well as to a certain topic within that theme.
The theme and topic are metadata which can be used to categorise the recordings.

Between the recording prompts, the participant is asked multiple questions about his or her background.
The metadata questions include dialect background, gender, native language, age, place of residence, birthplace, occupation and education. In this paper, we focus on the first four of these metadata types.

The dialect background question offers 20 options to choose from. In order to have fewer classes, we clustered these dialect regions into eight larger dialect groups, based on the information provided by The Institute for the Languages of Finland\footnote{\url{kotus.fi/en/on_language/dialects/finnish_dialects_7541}}. The dialect groups and their abbreviations used in this paper are:
\begin{enumerate}
    \item The Southwestern dialects (SW)
        \begin{itemize}
            \item Varsinais-Suomi
            \item Ahvenanmaa
        \end{itemize}
    \item The transitional dialects between the Southwestern and Häme dialects (TRAN)
        \begin{itemize}
            \item Uusimaa
            \item Satakunta
        \end{itemize}
    \item The Häme (Tavastian) dialects (HÄME)
        \begin{itemize}
            \item Pirkanmaa
            \item Häme
        \end{itemize}
    \item The dialects of South Ostrobothnia (Pohjanmaa) (SO)
        \begin{itemize}
            \item Etelä-Pohjanmaa
            \item Pohjanmaa
        \end{itemize}
    \item The dialects of Central and North Ostrobothnia (Pohjanmaa) (CNO)
        \begin{itemize}
            \item Keski-Pohjanmaa
            \item Pohjois-Pohjanmaa
        \end{itemize}
    \item The dialects of Peräpohjola (the Far North) (FN)
        \begin{itemize}
            \item Lappi
        \end{itemize}
    \item The Savo dialects (SAVO)
        \begin{itemize}
            \item Pohjois-Savo
            \item Etelä-Savo
            \item Kainuu
            \item Keski-Suomi
            \item Pohjois-Karjala
            \item Kymenlaakso
            \item Päijät-Häme
        \end{itemize}
    \item The Southeastern dialects and a few transitional dialects bordering on them (SE)
        \begin{itemize}
            \item Etelä-Karjala
        \end{itemize}
    \item Non-native Finnish speakers (NN)
\end{enumerate}

\subsection{Corpus statistics}
In the \textit{Lahjoita puhetta 2021} release, there are about 3600 hours of recordings in total, and over 20k different speakers. The median speaker donated 8 recordings, while the top donor donated 1039 recordings. The median duration of a recording is about 40 seconds and the longest are about 10 minutes.

Silent parts were trimmed from the beginnings and endings of the recordings using the \texttt{silence} effect of SoX\footnote{\url{sox.sourceforge.net}}, with a threshold of 0.5\% and duration of 0.05 seconds. After trimming, 
3270 hours remained, and the randomly selected recordings were sent to human transcribers. When we received the transcribed subset, there were 512 recordings that had empty transcriptions. Some of these were silent audio and some were left empty by mistake by the transcribers, but all 512 were discarded at this point. 
To verify the quality of the human transcriptions, 
we generated ASR transcriptions 
with a hybrid HMM/DNN (hidden Markov model / deep neural network) system trained on the previously existing colloquial Finnish speech data: DSPCON, FinDialogue, SPEECON (see the Introduction). The average WER (word error rate) was around 38\% and CER (character error rate) about 15\%. We then filtered out recordings for which both the WER and the CER were over 94\% in order to mitigate the chance of having low-quality samples in the ASR training corpus. From the set of about 100k transcribed recordings, 392 had WER and CER over the threshold and were excluded. Combined with the 512 empty-transcript recordings, these excluded 904 recordings were about 9.1 hours in duration.

\begin{table*}[htb]
    \caption{The sizes of the corpus and its subsets.}
    \label{tab:corpus_stats}
    \centering
    \begin{tabular}{cccc}
    \toprule
    \multicolumn{1}{c}{\textbf{Subset}} & \multicolumn{1}{c}{\textbf{\# of speakers}} & \multicolumn{1}{c}{\textbf{\# of recordings}} & \textbf{\# of hours} \\ \midrule
    Total original & 20890 & 218146 & 3604.8 \\
    Total usable & 20269 & 205962 & 3229.8 \\ \midrule
    train transcribed & 17821 & 98606 & 1601.5 \\
    train untranscribed & 18825 & 105380 & 1597.1 \\ 
    train transcribed 100h & 1129 & 6229 & 103.5 \\ 
    dev & 103 & 703 & 10.5 \\ 
    test & 103 & 690 & 10.4 \\
    test multi-transcriber & 57 & 58 & 1.0 \\
    test multi-transcriber speakers & 57 & 583 & 10.2 \\  
    \bottomrule
    \end{tabular}
\end{table*}

We sampled a 10-hour test set and 10-hour development set from the transcribed speech data, each including at least ten minutes of speech for each metadata class in each of the five metadata domains. The gender ratio has also been debiased, so that the dev and test sets have over 40\% male speakers although the training set has just over 20\%. As a second test dataset, we used a 1-hour set that was transcribed by four different transcribers, which includes 58 recordings from 57 speakers. 
If we add all recordings by those 57 speakers to this subset, we get a 10-hour test set, that we call "test multi-transcriber speakers" in Table \ref{tab:corpus_stats}.
The rest of the transcribed speech is used as training data. The train, dev and test sets have no overlap of speakers. There are still recordings that are by the speakers of the dev or test sets but which are not transcribed. These are left unused, leaving about 3230 hours in the complete dataset that we use. Table \ref{tab:corpus_stats} lists the sizes of the corpus subsets.

\begin{figure}[htb]
    \centering
    \includegraphics[height=13cm]{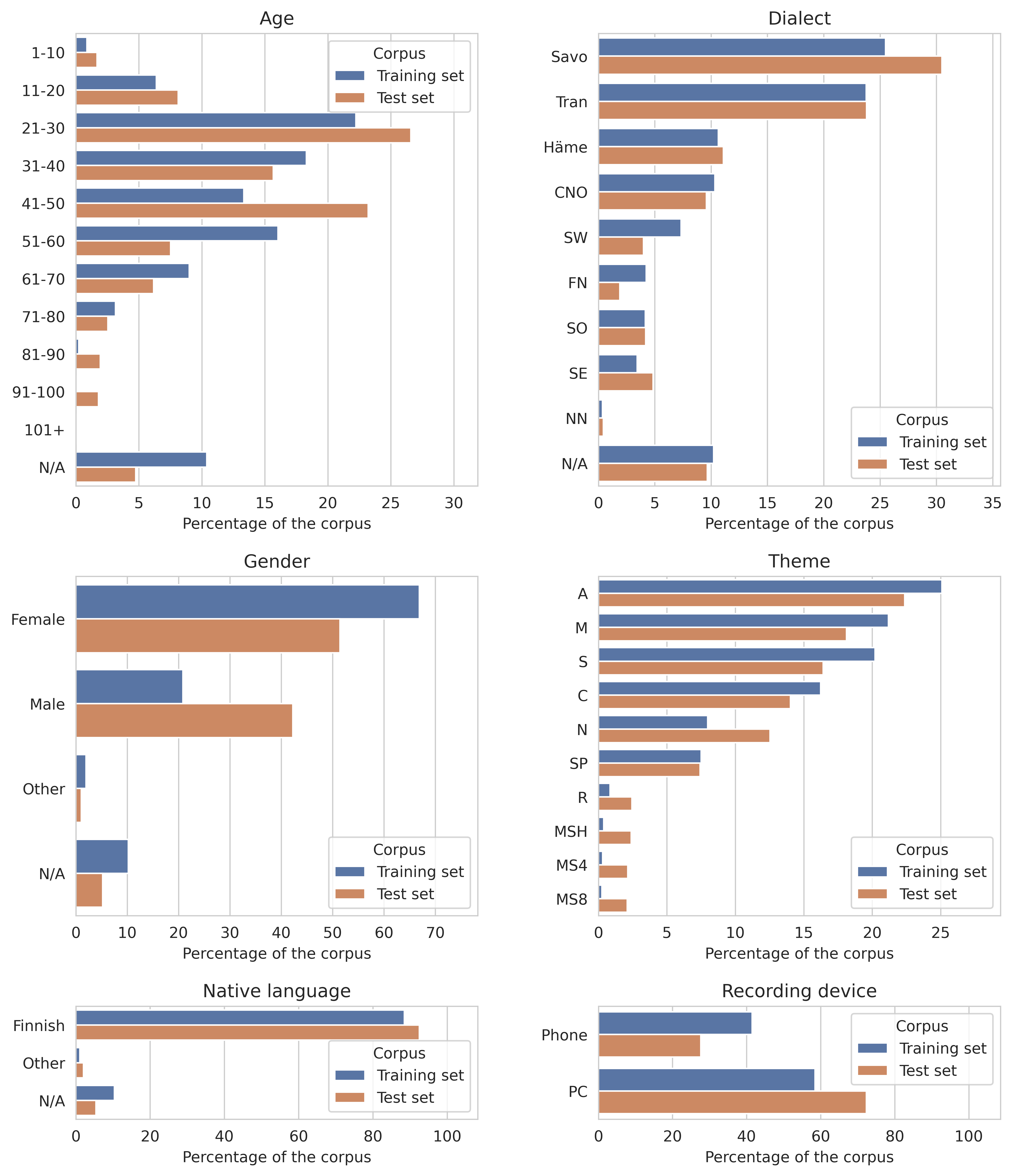}
    \caption{The distribution of the speaker metadata in the corpus. The “training set” includes both the "train transcribed" and "train untranscribed" described in Table \ref{tab:corpus_stats}. “N/A” means the user has not answered to the question about his or her background, or has given multiple contradicting answers.}
    \label{fig:metadata_stats}
\end{figure}

Figure \ref{fig:metadata_stats} presents the amount of speech for each metadata type as a portion of the whole training set (both transcribed and untranscribed pooled together) and the 10-hour main test set. As the transcribed training set is a 1600-hour random sample of the whole data set, which has about 3199 hours out of the complete 3230 hours, the training data accurately represents the overall distribution of the whole data set. We can note that the corpus has varying amounts of speech from the different metadata classes. Younger than 11-year old children have donated some but a relatively small amount, as have older than 80-year-old people. Of the dialects, Savo and Tran have most data, roughly a quarter each. Women have donated significantly more than men: over three times as much. Four themes seem to have a low amount of speech: Rated R and the three Media skills themes, as they were added only when the official marketing campaign had already ended. In all metadata domains, the test set was smoothed to have at least 10 minutes of speech from each metadata class, visible in the figure.

\begin{figure}[htb]
    \centering
    \includegraphics[height=8cm]{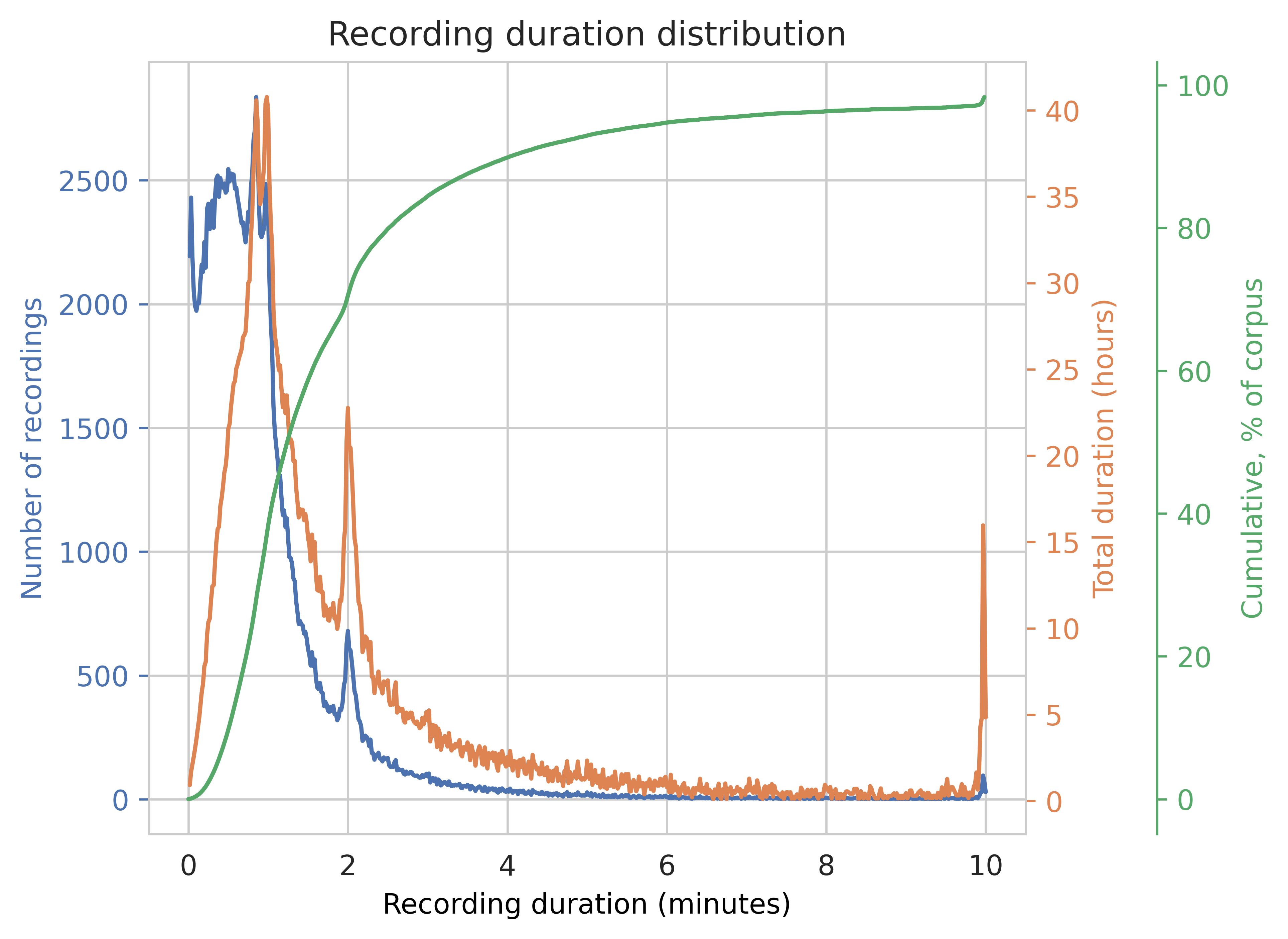}
    \caption{The recording length distribution. The recording durations are pooled to 1-second bins to generate this figure.}
    \label{fig:duration_stats}
\end{figure}

Figure \ref{fig:duration_stats} displays the distribution of the recording lengths. The majority of the recordings are less than two minutes, but longer recordings are not uncommon. There are spikes at the 2-minute mark and the 10-minute mark. The spike at 10 minutes was effected by the limit of the duration of recordings: those that would have spoken for longer were cut at 10 minutes. The other spike, at two minutes, corresponds to the duration of a video clip that was played for the user in one topic. The theme was “Summer”, and in this topic the user was asked to describe what is happening in the video clip to an alien while the video displayed sceneries of Finnish summer pastime activities. 

\section{Annotation procedure}
Because a high-quality manual transcription of 1600 hours of spontaneous speech is a significant investment, we made an effort to develop a careful process described in detail in this section. 
The aim was exact transcription, which included not only the verbal content of the speech but also full words, repetitions, hesitations, partially pronounced or only partially audible words, and non-verbal communication such as laughs, growls, and coughs. The guidelines that were given to the transcribers are reproduced in Appendix \ref{ap:transcriber_instructions}.

\subsection{First phase: annotator selection}
To choose the best transcriber companies, we ran a pilot transcription competition, where we shared a 20-hour subset of the data with all candidates along with the carefully constructed annotation instructions. The datasets consisted of 19 hours of randomly selected data per participant mixed with a common one-hour evaluation set (the composition of the data was not disclosed to the companies). After the competitors submitted their transcripts, we evaluated them automatically and manually using the overlapping one-hour set to determine the quality of their work as well as an hour of random samples from the non-overlapping parts to verify the automated comparisons manually.

The automatic evaluation focused on comparing the transcripts of different annotators with each other and with multiple ASR systems. Our goal was to find the best transcribers, so we used standard ASR metrics like word error rate (WER) and character error rate (CER) to compare the performance of the annotators. Specifically, we treated one annotator as an ASR model and compared the produced transcription with another annotator's text, which was considered the gold standard. This allowed us to create rankings from the gold standard annotator's perspective, assuming that lower error rates correspond to better transcription quality.
During these analyses, we ignored the non-word symbols, as they were annotated with considerable discrepancies by different annotators.


\begin{table}
    \caption{Pairwise comparison between transcribers (T) using the word and character level edit distances.}
    \label{tab:inter_annot_er}
    \begin{tabular}{c|ccc}
        \toprule
        \multicolumn{4}{c}{\textbf{Word level comparisons}} \\
        \midrule
        \textbf{Transcriber} & T2 & T3 & T4 \\
        \midrule
        T1&~19.5\%~&~19.8\%~&~20.5\%\\
        T2&~-~&~13.6\%~&~15.6\%\\
        T3&~-~&~-~&~16.0\%\\
        \bottomrule
    \end{tabular} \hfill
    \begin{tabular}{c|ccc}
        \toprule
        \multicolumn{4}{c}{\textbf{Character level comparisons}} \\
        \midrule
        \textbf{Transcriber} & T2 & T3 & T4 \\
        \midrule
        T1&~6.3\%~&~5.8\%~&~6.1\%\\
        T2&~-~&~4.7\%~&~5.4\%\\
        T3&~-~&~-~&~4.9\%\\
        \bottomrule
    \end{tabular}
\end{table}

The inter-annotator disagreements in terms of WER and CER were generally high due to the nature of the data, see Table~\ref{tab:inter_annot_er}. Still, we can observe considerable differences. These metrics allowed us to create rankings per annotator. Fortunately, we only wanted to ensure the high quality of the transliteration, so we did not have to use complex methods (like the Borda count etc.) to produce a complete order. In the end, we opted for a straightforward scheme to aggregate the individual preference orders by simply eliminating the worst in each round until we get the desired number of annotators.

Looking at the values in Table~\ref{tab:inter_annot_er}, we can see that T1 had the highest disagreement with the others, both in terms of WER and CER. The transcription quality was also substantiated by manually inspecting 1-hour random samples from each candidate. Thus T1 was the first to be eliminated. Of the remaining annotators, T2 and T3 disagree most with T4. Nevertheless, the differences between these three annotators were relatively small, so in the end, we opted to accept all three in this round of selection.

\begin{table}
    \caption{Pairwise comparison between transcribers (T)  and ASR systems using the word and character level edit distances.}
    \label{tab:compare_annot_to_asr}
    \begin{tabular}{c|cc}
        \toprule
        \multicolumn{3}{c}{\textbf{Word level comparisons}}\\
        \midrule
        \textbf{Transcriber} & \textbf{Hybrid} & \textbf{E2E} \\
        \midrule
        T1&~33.56\%~&~33.65\%~\\
        T2&~28.02\%~&~27.33\%~\\
        T3&~29.04\%~&~28.83\%~\\
        T4&~29.87\%~&~29.87\%~\\
        \bottomrule
    \end{tabular} \hfill
    \begin{tabular}{c|ccc}
        \toprule
        \multicolumn{3}{c}{\textbf{Character level comparisons}}\\
        \midrule
        \textbf{Transcriber} & \textbf{Hybrid} & \textbf{E2E} \\
        \midrule
        T1&~11.95\%~&~10.12\%~\\
        T2&~10.14\%~&~8.59\%~\\
        T3&~10.69\%~&~8.89\%~\\
        T4&~10.93\%~&~9.15\%~\\
        \bottomrule
    \end{tabular}
\end{table}

Next, we repeated the experiments, but this time, we compared the transcripts with ASR outputs. Two models were selected for this purpose, a hybrid HMM/DNN, and a Wav2Vec2-based \citep{w2v2} end-to-end network.
The hybrid HMM/DNN system was trained on the existing spontaneous colloquial Finnish speech datasets: DSPCON, FINDialogue and SPEECON (spontaneous part), totalling about 37 hours. The 1st pass n-gram LM and 2nd pass RNN LM are trained on the WEBCON \citep{enarvi2018modeling} corpus and the speech transcripts, in total about 76 million words. For the end-to-end model, we decided to utilise the publicly available multilingual \textit{Wav2Vec2 Large} model pre-trained on 100K hours of the VoxPopuli dataset  \citep{wang2021voxpopuli}. The model was fine-tuned on the same 37-hour colloquial Finnish corpus used to train the hybrid system.

Comparing with ASR models reaffirmed our previous findings (Table~\ref{tab:compare_annot_to_asr}). We can see that comparing the ASR models with T1 leads to the highest error rates. An interesting observation is that both models seem to
favour T2, yielding the lowest error rates, followed by T3 and T4. 

Lastly, we also validated the conclusions of all automatic experiments by manually checking the utterances 
with the largest differences (revealed by the previous examinations). 
The manual inspection revealed that T4 had transcribed files mostly correctly, but they often used the formally correct spelling instead of writing the verbatim spoken version. This resulted in slightly higher error rates compared with T2 and T3. Comparing T2 and T3  we saw that the latter skipped the extremely noisy part of an utterance, resulting in T2 being selected as the most diligent annotator. 

Combining all observations, we concluded that T2, T3, and T4 are all capable of creating sufficiently high-quality transcripts, so we continued to work with them to transliterate a large portion of the collected corpus.

\subsection{Second phase: quality control} 
After the initial selection phase, we continued to utilise our ASR models to perform automatic quality control checks. Our goal was to highlight recordings with unusual error rates and investigate the reasons for the high errors. In practice, once we received the transcriptions from the companies, we applied the same ASR models as in the phase one to get the WER and CER for each utterance. To avoid unnecessary checks, we only selected files with a high WER and CER compared with both models. 

Our manual examinations revealed several problems that we could address during the annotation process. One of the primary issues that we managed to identify was a mismatch between the transcription and the audio files (approx. 20 transcripts had been assigned to the wrong recording). Naturally, with the help of the annotators, we could fix this problem quickly. The second source of the high ASR error rates was the presence of extreme noises, which made it hard for the ASR systems to recognise the speech. We kept these noisy recordings in the corpus to enable the building of noise-robust models.

\begin{figure}
    \centering
    \includegraphics[width=.45\linewidth]{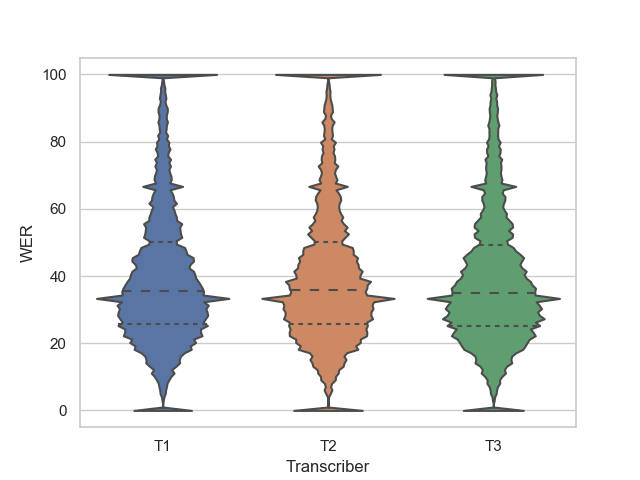} \includegraphics[width=.45\linewidth]{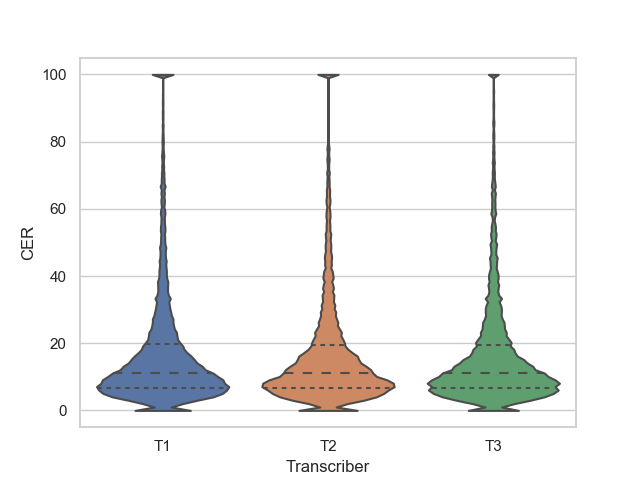}
    \caption{The distribution of word-level (left) and character-level (right) error rates per annotators on the transcribed dataset. Note: utterances with more than 100\% errors were pooled together for this visualisation. Note also that the transcribers' ids of the second phase do not match to the first phase.}
    \label{fig:error_stats}
\end{figure}

Figure~\ref{fig:error_stats} depicts the error rates of the hybrid ASR model for each transcriber. Note that due to legal constraints, we were unable to match the transcribers' ids used here to those in the first phase. Thus we could not analyse how their performance changed on the large dataset. Overall, we can see that the distributions are quite similar, meaning that from the ASR model's viewpoint, they were equally good at providing the gold standard texts. We can see that there is a considerable amount of utterances with more than 100\% WER, but overall, the vast majority of recordings are recognisable with less than 50\% error. The CER statistics further reassured us that the transcription is high quality; more than 75\% of the utterances had a CER below 20\%. The high errors could be explained by the discovered problems (noise, low volume, speaking far from the microphone). 

\section{ASR experiments and results}
In this section several ASR experiments with various architectures are presented.
The goal of the ASR experiments is first to establish that the transcribed Lahjoita puhetta data is useful for creating ASR systems, and then to provide baseline results and recipe starting points for a few different ASR techniques. The trained ASR systems are also used to provide both time alignments of the manually transcribed part, as well as ASR decoding outputs for the untranscribed part, which can later be used for indexing, searching, or statistical studies on the data, as attested by for example \citet{carrive2021avanalysis}.

One initial difficulty in using the transcribed Lahjoita puhetta data for ASR is that many of the recordings are longer in duration than is ideal for many speech recognition methods. Bootstrapping alignments for long recordings is more difficult. Long recordings exacerbate the vanishing gradient problem and they also present practical issues related to memory consumption~ \citep{narayanan-longform-recognition-e2e}. In these experiments, we are able to bootstrap alignments and create shorter segmentations for different systems by starting from simple monophone HMM/GMM (Gaussian mixture model) systems trained on the shortest utterances.

It is good to note that as Finnish is an agglutinative language, the WER results are not directly comparable to those of, say, English. \citet{hirsimaki2006unlimited} found that as one long Finnish word corresponds to several English words the WER becomes multiplied. For this reason, we report also the CER results, which do not have this problem. Furthermore, some previous works \citep{enarvi2017automatic} have used normalisation of colloquial Finnish words in order to mitigate the effect of various spelling variations on the WER results. However, this method is partly manual and thus not easily scalable to large corpora, and we did not use such normalisation.
Additionally, the transcripts contain special markers (e.g. for noise and pauses) and some decisions should be made about them in speech recognition: either to predict them, or to simply discard them. We opted for the latter. Before calculating the WER and CER, we removed all the special tokens, such as ".laugh" as well as the dash symbols "-" that indicate dysfluencies in speech, for example false starts ("predi- presidentti" was changed to "predi presidentti"). 

\subsection{Hybrid HMM/DNN ASR systems} \label{sec:hybrid}
We trained some baseline HMM/DNN ASR systems using the Kaldi \citep{Povey_ASRU2011} toolkit.
In the first phase, we trained two models using mostly standard Kaldi recipes without hyperparameter tuning, one with a 100-hour subset (denoted as initial-100h-TDNN) and another with the complete transcribed training corpus (initial-1600h-TDNN). To train the HMM/GMM system for monophones and triphones, we used the Kaldi WSJ recipe.
This recipe trains the initial monophone model on the shortest utterances in the data, which helps in bootstrapping the alignments. As a deviation from the standard WSJ recipe, we trained the final triphone system using the discriminative, MMI \citep{bahl1986maximum} training criterion, which is available as an optional addition in the WSJ recipe. The time-delay neural network (TDNN) \citep{waibel1989phoneme,peddinti2015time} models were trained using the HMM/GMM alignments. The TDNN architecture and other hyperparameters were adopted from the Switchboard recipe, since this trains a larger neural network, more suitable for the large training corpus. The TDNN has 15 layers with a dimension of 1536 and a bottleneck dimension of 160. In total the TDNN has about 17M parameters.

Using the SRILM \citep{stolcke2002srilm} toolkit, we trained 4-gram language models (LMs) on the 100-hour transcriptions, the whole 1600h training corpus transcriptions, as well as on pooled corpora of other available colloquial Finnish text corpora, namely the WEBCON corpus and the DSPCON transcriptions, and the Lahjoita puhetta (LP) transcriptions (both 100h and 1600h separately). The systems that utilised also the external language modelling data are marked with “ext. LM data” in Table \ref{tab:hybrid_res}. 
\begin{table}[ht!]
\centering
\caption{Sizes of the language models and their training corpora. A 4-gram LM includes also unigrams, bigrams and trigrams, but we list only the number of 4-grams for comparison here.} \label{tab:lm_size}
\begin{tabular}{cccc}
    \toprule
    \textbf{Corpus} & \textbf{\# of tokens} & \textbf{\# of types} & \textbf{\# of 4-grams}\\
    \midrule
    LP 100h transcriptions & 898 700 & 20 700 & 24 800\\
    LP 1600h transcriptions & 14 216 500 & 36 200 & 1 865 400\\
    \midrule
    WEBCON+DSPCON+LP100h & 126 078 900 & 42 700 & 7 379 600\\
    WEBCON+DSPCON+LP1600h & 138 991 000 & 45 600 & 8 305 900 \\
    \bottomrule
\end{tabular}
\end{table}
We used the Morfessor \citep{creutz2002unsupervised,creutz2007unsupervised} toolkit to segment words into subword units. We trained the morfessor model using the same LP transcriptions appended with the WEBCON and DSPCON corpora as for the large LMs, with a corpus weight of 0.05. The resulting sizes of the LMs and their training corpora are listed in Table~\ref{tab:lm_size}. We also tried to use a word vocabulary, but subword units yielded better results. For example, the word-based initial-1600h-TDNN system got a WER of 25.12 on the test set, compared with 24.00 using subword units. The sizes of the training corpora are listed in Table \ref{tab:lm_size}. For more details about the language models, see the published recipes.

We used the initial-1600h-TDNN to segment the training data, so the data could be used for training the E2E ASR systems. The initial-100h-TDNN with large LM was used to generate transcriptions for the rest of the training corpus, which we then used for training the topic and dialect classification systems (see Section \ref{sec:metadata_classification}). 

After training the initial ASR systems, we made some simple hyperparameter tuning for the HMM/GMM system to get an idea of how much room for improvement there is, compared with the Kaldi WSJ recipe. The tuning experiments focused mainly on increasing the number of parameters of the GMMs. By increasing the number of Gaussians from 4200 (in the WSJ recipe) to 14000, and the number of leaves per Gaussian from 40000 to 200000, the penultimate, speaker-adaptive triphone system WER on the development set decreased from 42.86 to 39.71. Training the MMI triphone system on top of the alignments from these systems, the WERs decreased to 37.08 and 35.36, respectively for the smaller and larger GMM/HMM system. Finally, training the TDNN system on top of these MMI triphone models, the word error rates dropped to 22.09 (smaller GMM/HMM) and 21.98 (larger GMM/HMM) for the dev set and 24.00/23.88 for the test set.

\begin{table}[ht!]
\centering
\caption{Error rates of various 
ASR systems. Larger LMs were trained on external LM data, namely the WEBCON corpus and the DSPCON transcriptions, in addition to the AM training set (either 100h or 1600h) transcriptions. All HMM/DNN system LMs are subword-based 4-gram models. The Wav2Vec2+CTC system uses a word-based 4-gram language model trained on the 1600h LP transcripts and the external data.}
\begin{tabular}{c p{4.3cm} p{0.7cm} p{0.7cm}  p{0.7cm} p{0.7cm}}
    \toprule
    &&\multicolumn{2}{c}{\textbf{Dev set}}&\multicolumn{2}{c}{\textbf{Test set}}\\ \cmidrule(lr){3-4}\cmidrule(lr){5-6}
    \textbf{Training set}&\textbf{System details}&\textbf{WER} & \textbf{CER} & \textbf{WER} & \textbf{CER} \\
    \midrule
    \multirow{6}{*}{100h}&initial-GMM & 44.80 & 18.43 & 48.94 & 20.94 \\
    &initial-TDNN & 29.16 & 9.01 & 32.58 & 11.04 \\
    &\quad \quad + ext. LM data & 26.88 & 8.46 & 30.48 & 10.49 \\
    &semisup-TDNN + ext. LM data & 25.37 & 7.89 & 28.16 & 9.78 \\
    &Wav2Vec2+CTC (no LM) & 22.50 & 6.08 & 24.03 & 7.02 \\
    &Wav2Vec2+CTC + ext. LM data & 20.34 & 5.83 & 21.75 & 6.80 \\
    \midrule
    \multirow{6}{*}{1600h}&initial-GMM & 37.08 & 15.61 & 40.87 & 17.19 \\
    &large-GMM & 35.36 &14.34 & 38.99 & 16.33 \\ 
    &initial-TDNN & 22.09 & 6.52 & 24.00 & 7.64 \\
    &TDNN (large-GMM alignments) & 21.98 & 6.47 & 23.88 & 7.59\\
    & \quad \quad + ext. LM data & 21.77 & 6.40 & 23.82 & 7.52 \\
    & AED                    & 28.80 & 12.15 & 34.87 & 17.04 \\
    \bottomrule
\end{tabular}
\label{tab:hybrid_res}
\end{table}

Decoding with a large language model trained on external data brings additional improvement compared with the LM trained on 100h transcriptions (see the second and third row in Table \ref{tab:hybrid_res}). However, the 1600h transcriptions seem to be enough to train a decent language model, and adding external data only brings a small improvement in WER and CER results (see the last two rows in Table \ref{tab:hybrid_res}). It is good to note, however, that the external text data is not exactly in the same domain as the test corpus, although it is colloquial in style. 

Additionally, we wanted to demonstrate that the sizeable untranscribed portion of the corpus can be leveraged via semi-supervised training. For this experiment, we choose the approach presented in~ \citep{Manohar2018SemiSupervisedTO}. To demonstrate that the recordings without annotations could be used for improving the ASR systems, we started the semi-supervised training by generating transcriptions of the additional data with the \textit{initial-100h-TDNN}. Afterwards, we pooled the self-supervised portion (approx. 1587 hours) and the 100h set for the model training. The resulting model (\textit{semisup-100h-model}) had the same architecture as the \textit{initial-100h-TDNN} to ensure a fair comparison. From the achieved results (see Table~\ref{tab:hybrid_res}), we can conclude that the additional unsupervised data is indeed valuable, the error rates dropped significantly.  On the other hand, we can also see that having more, accurately transcribed data is far more beneficial. The \textit{initial-1600h-TDNN} outperforms the semi-supervised system by a large margin, and the hyperparameter tuning offers some additional improvements. 
    
\subsection{AED ASR systems}
Various end-to-end ASR approaches, such as Connectionist Temporal Classification (CTC)~\citep{graves2006connectionist}, the Recurrent Neural Network Transducer (RNN-T)~\citep{ctc-rnnt-asr}, and Attention-based Encoder-Decoder (AED)~\citep{bahdanau2016attention,chan2016las} models became popular in the 2010s, both in research as well as industrial applications. We train AED models on the transcribed data to serve as end-to-end baselines. Our AED models are trained with the SpeechBrain toolkit~\citep{speechbrain}. They consist of a stack of convolution, recurrent, and feed-forward layers in the encoder, a location-and-content aware attention mechanism, and recurrent layers in the decoder with altogether $\approx 28$M parameters. The inputs are log-Mel-filterbank-energies and for each output step the network computes a distribution over a vocabulary of 1750 SentencePiece subword units. We trained with dynamic batching, targeting 50 seconds of audio per batch altogether, for 100 nominal epochs of 10000 updates each. For the first 20 nominal epochs the encoder learning was aided by using an additional multi-task CTC loss~\citep{jointattentionctc}. We do not use any external language with our AED system, making it fully end-to-end. For further details we refer to the published recipe.

End-to-end models seem to have difficulties with long-form speech, both in learning as well as in generalising~ \citep{chiu-e2e-longform-cmp-19,narayanan-longform-recognition-e2e}. Our preliminary experiments with AED systems showed similar issues. Models would not converge with full length utterances. Via segmentations produced with the HMM-based ASR systems, we split the data into shorter utterances. Training converges well on short (up to 10 second) segments and slightly slower on medium length (up to 50 second) segments. Decoding an ad-hoc segmented version of the development set yields a WER of $\approx 22\%$ on both models. However, on the official development set, which has longer utterances, both models have pathological behaviour on a minority of utterances, which increased the error rate considerably. Similar to reports by~\citet{keung2020attentional}, our models produce echographic output, i.e. the model repeats a single token or in some cases a long sequence of tokens. The model trained on medium length segments suffers less, so we choose it as our final baseline. Additionally, we implement a simple post-processing filter where we allow repetitions to produce in total a maximum of five tokens. On the development set, this modifies 70 transcripts. This reduces the WER from $45.82\%$ to $28.80\%$ - echographic transcripts account for a significant amount of errors. Listening to the utterances which produced echographic output reveals that these utterances are long, in some cases noisy, and in some cases contain long pauses. Despite the post-processing, our AED baselines fall behind their HMM/DNN counterparts in performance in table~\ref{tab:hybrid_res}. Due to the initial difficulties with long-form speech, we did not make a system for the 100h subset.

\subsection{Pretrained Wav2Vec2 fine-tuned with CTC}

\textit{Wav2Vec2} \citep{w2v2} is a self-supervised framework which learns deep acoustic representations by leveraging large amounts of unlabelled acoustic data. After pre-training on untranscribed speech, the model can be fine-tuned on labelled acoustic data for a downstream task, such as ASR. Fine-tuning for the ASR starts with adding a randomly initialised classification layer on top of the model with classes representing the characters of the target language alphabet and a word boundary token. The model is then optimised with a CTC loss.

In this work, we experimented with a \textit{Wav2Vec2 Large} model (317M parameters) pre-trained on the multilingual VoxPopuli \citep{wang2021voxpopuli} corpus. The corpus is composed of 100K hours of untranscribed European Parliament plenary session recordings in 23 languages, including 4.4K hours of Finnish speech. We fine-tuned this model with CTC on the 100-hour subset for 80 epochs with an effective batch size of 48 and a learning rate of 5e-4. We used full length utterances with durations up to 50 seconds and the segmented recordings for the rest of the training data. We also tried to fine-tune the model on the 1600h set, but it took too much time on our hardware, so we left fine-tuning on the full training set to future work.

The fine-tuned model (see \textit{Wav2Vec2+CTC (no LM)} in Table \ref{tab:hybrid_res}) achieved WER of 22.50\% and 24.03\% and CER of 6.08\% and 7.02\% on the development and the test set, respectively. We also incorporated an external language model in order to further improve the model performance. The LM was trained on the 1600h LP transcriptions and external (WEBCON and DSPCON) data. The dataset included about 84M word tokens and 2.6M word types, and the LM included 2.5M 4-grams. With a word-level 4-gram LM (see \textit{Wav2Vec2+CTC + ext. LM data} in Table \ref{tab:hybrid_res}), the word and the character error rates dropped to 20.34/5.83\% on the development set and 21.75/6.80\% on the test set. In addition, we plan to incorporate the subword-based LM in future experiments, since it provided an improvement in WER compared to word-based LM for some HMM/DNN ASR systems.

\subsection{Analysis of ASR accuracy w.r.t speaker metadata}
The rich metadata of Lahjoita puhetta (see for example Figure \ref{fig:metadata_stats}) allows us to examine differences in the ASR accuracy between speech from different groups of people, different recording devices and on different topics. A basic assumption is that the more training data there are from a specific group the better the speech recognition results are for this group. This means the correlation between the amount of training data and WER result should be negative. 

Dividing the 10-hour test corpus into each metadata class yields quite small subsets. To get a larger test corpus for each metadata class, we decoded the rest of the transcribed data set using the initial-100h-TDNN ASR system with external LM data (see Section \ref{sec:hybrid}). In this case there is overlap between training and test corpus speakers, although no overlap between the recordings. The average WER for this large set was 26.13\% which is somewhat lower than for the fully independent 10-hour test set (see Table \ref{tab:hybrid_res}).

\begin{table}[htb]
    \centering
    \caption{Pearson product-moment correlation coefficients between the WER and the total duration of speech in the training corpus. }
    \label{tab:wer_traindata_corr}
    \begin{tabular}{cc}
    \toprule
    \textbf{Metadata type} & \textbf{Correlation coefficient} \\
    \midrule
    Age & -0.685 \\
    Gender  &  -0.618 \\
    Dialect & -0.267 \\
    Theme &  -0.829 \\
    Device & 1 \\
    \bottomrule
    \end{tabular}
\end{table}

Table \ref{tab:wer_traindata_corr} lists the Pearson product-moment correlation coefficients between the WER and the amount of training data, for each metadata type. Age, gender and theme have the expected results, with quite a strong negative correlation. Dialect has a weaker negative correlation, and for the recording device (phone vs. PC), WER correlates positively with the amount of training data.  

\begin{figure}[!ht]
    \centering
    \includegraphics[width=0.8\linewidth]{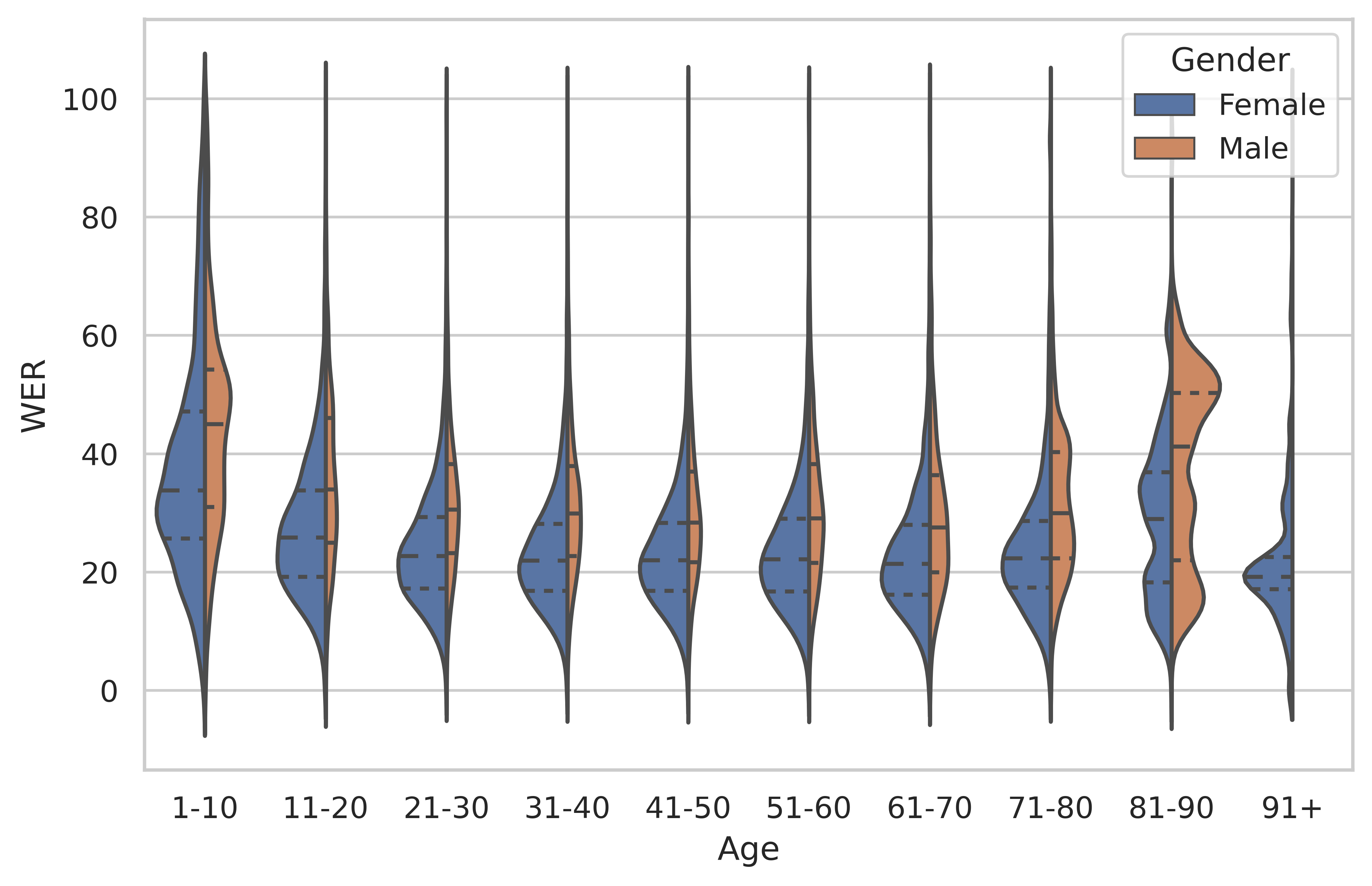} 
    \caption{The distribution of WERs in the test set w.r.t. the age and gender of the speaker.}
    \label{fig:age_and_gender_wer}
\end{figure}

\begin{figure}[!ht]
    \centering
    \includegraphics[width=0.8\linewidth]{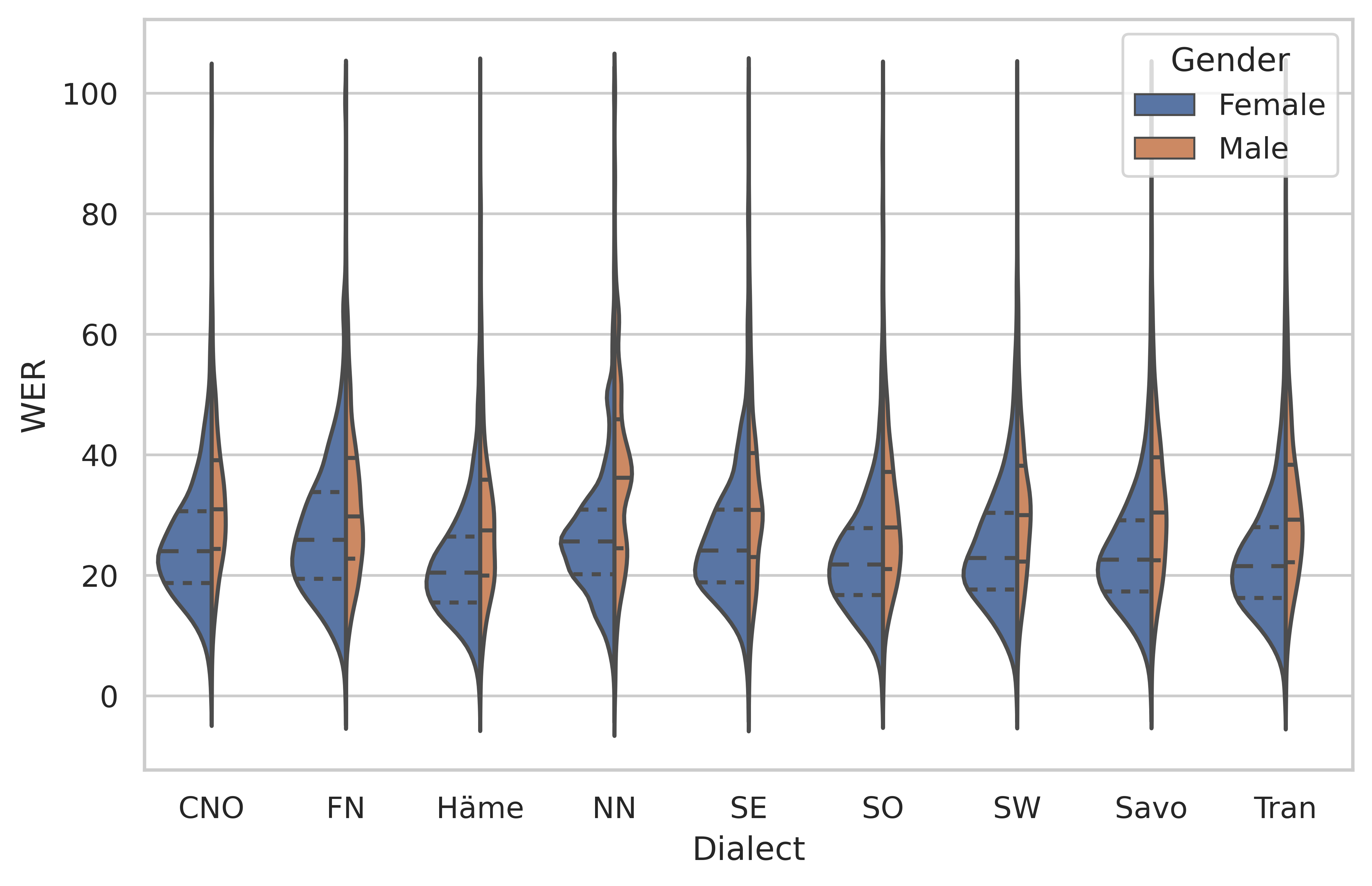} 
    \caption{The distribution of WERs in the test set w.r.t. the dialect and gender of the speaker.}
    \label{fig:region_and_gender_wer}
\end{figure}

Figures~\ref{fig:age_and_gender_wer} and \ref{fig:region_and_gender_wer} enable a more detailed analysis of the results for the metadata groups of gender, age and dialect. The difference between the number of males and females in the training data is large, which clearly affects the ASR results. The average WER for females, 24.12\%, is below the overall average (26.13\%) while for males the average WER is well above: 31.78\%.
Similarly, the number of recordings on a topic in the training corpus correlates with the speech recognition accuracy. The experiments presented in Section \ref{sec:metadata_classification} actually verify that the transcriptions were more useful than the full audio in the topic classification task. Therefore, although the gender difference is probably due to acoustic model training, the differences of ASR results of different topics is more likely to be due to language model training. 

The dialect seems to have a relatively weaker effect on the speech recognition results than than gender and topic. The large dialect groups, Savo and Tran, do not have significantly better results than the average: the WERs are 25.99 and 25.13 respectively. This can be seen also from the correlation coefficient in Table \ref{tab:wer_traindata_corr}. The exception is the group of non-native Finnish speakers, which has a high WER of 30.97\%. 

From these experiments, the reasons for the differences in the WER results are not entirely clear. In general, a larger relative share in the training corpus results in better ASR performance, but other factors presumably affect the results too. For example, there is fewer training data for the speech of young children which might be one reason for the relatively poor ASR performance, but children probably speak less clear Finnish than adults, which also makes speech recognition more difficult. This could apply also to other groups, such as non-native Finnish speakers. Furthermore, speech recorded on smart phones has better ASR accuracy (WER: 24.57\%) than speech recorded on a computer (WER: 27.27\%) even though there are fewer phone recordings than computer recordings. We speculate that the reason for this is that phones are better than computers, on average, at recording speech.

In recent years, a new research area has emerged that investigates the discriminatory performance of AI systems and its causes~ \citep{hovy-spruit-2016-social,garnerin-etal-2021-investigating}. In the ASR field, traditional metrics like the aggregated WER and CER are used to measure the overall performance of the models. As we have noted in this section, these metrics can hide biases that a model develops during training. To build an excellent general ASR system, we ought to mitigate the risk of the system having a systematically
worse recognition rate for any speaker category (for example, gender, age, or dialect). The results analysed in this section can point us to the weaknesses of the system and aid us in future de-biasing efforts.

\section{Gender, age, dialect, and topic classification} \label{sec:metadata_classification}

The big strengths of the Lahjoita puhetta dataset, besides its size, are the variety of speakers and the rich metadata provided by them. Using this metadata, we can build various metadata classifiers, which can later be used in different applications, such as: filling the missing metadata, verifying the correctness of the available metadata, enhancing the speech processing applications with speaker information, and bias detection. For that purpose, we built and benchmarked baseline models for gender, age, dialect, and topic classification. 

The models are built using a 5-layer TDNN with dilated connections, followed by statistical pooling and two linear layers. This is similar to the x-vector models  \citep{snyder2018x}. We will call this part audio encoder. For the dialect and topic classification tasks, besides the models trained on audio-only, we additionally trained models that utilise the available transcripts. We did that using an additional text encoder. In the text encoder, word embeddings are extracted using the FinBERT model  \citep{virtanen2019multilingual} and processed through a bi-directional long short-term memory (BLSTM) network  \citep{hochreiter1997long}. In the last stage, the outputs of the audio and text encoders are concatenated and passed through a softmax function which produces class probabilities.

As input features, we extracted logarithmic-Mel-filterbanks with 40 filters, using 25ms window with a stride of 10ms. To improve the signal-to-noise ratio, we applied mean normalisation to each sample.

The hyperparameters for the audio encoder are given in Table \ref{tab:audio_enc_params}. The text encoder is a 2-layer BLSTM with an input size of 768 and an output of 512. As optimiser, we used Adam  \citep{kingma2014adam}, with a learning rate of 1e-4 and a cross entropy loss.

\begin{table}[htb]
    \centering
    \caption{Hyperparameters of the audio encoder.}
    \label{tab:audio_enc_params}
    \begin{tabular}{ccccc}
    \toprule
    \textbf{Layer} & \textbf{Input size} & \textbf{Output size} & \textbf{Context} & \textbf{Dilation} \\
    \midrule
    TDNN 1 & 40 & 512 & 5 & 1 \\
    TDNN 2 & 512 & 512 & 3 & 2 \\
    TDNN 3 & 512 & 512 & 3 & 3 \\
    TDNN 4 & 512 & 512 & 1 & 1 \\
    TDNN 5 & 512 & 1500 & 1 & 1 \\
    \midrule
    Statistical pooling & 1500 & 3000 & / & / \\
    Linear & 3000 & 512 & / & / \\
    \bottomrule
    \end{tabular}
\end{table}

\subsection{Gender classification}
Gender information plays an important role in many applications, from speech processing  \citep{abdulla2001improving} to bias detection  \citep{park2018reducing}. Thus, having a good gender classifier can help us enhance the speech processing models, as well as aid us in detecting the biases related to gender, that those models may contain. For that purpose, we built two gender classifiers, using different segment lengths.

The audio samples used to train the first model are cut to 50 seconds. The reason for not using the whole audio samples is that some of them might be too long to process. Additionally, the 50 seconds limit of the audio should contain a sufficient amount of information for the model to learn the task.

The gender classification models usually work with small (few seconds) audio segments, whereas the average length of the audio samples in our dataset is about 40 seconds. To make our model more comparable to the others, we constructed another model that uses audio segments up to 3 seconds. This choice is expected to degrade the performance of the model but will make it more reusable to other applications, where long segments are not available.

\begin{table}[htb]
    \centering
    \caption{Accuracy of the models on the gender classification task.}
    \label{tab:gender_results}
    \begin{tabular}{ccc}
    \toprule
    \textbf{Model} & \textbf{Test} & \textbf{Multi-transcriber test} \\
    \midrule
    3 seconds model & 90.03 & 99.59 \\
    50 seconds model & 92.65 & 99.59 \\
    \bottomrule
    \end{tabular}
\end{table}

In Table \ref{tab:gender_results}, we can see how both models performed in terms of accuracy on the test sets (see Table \ref{tab:corpus_stats} for set descriptions). From the results, we can observe that on the test set, the model using up to 50 second segments performs slightly better than the one using 3 second segments. This is expected, considering that longer segments contain more information. On the multi-transcriber test set, on the other hand, both models perform equally well, achieving almost perfect accuracy score. The significant difference in performance between both test sets could be attributed to the disproportion between male and female speakers. The multi-transcriber test set has many more female speakers than male, and as we will see later, the system is better at detecting the female speakers.

\subsection{Age classification}
Like gender, age information can also be beneficial in many areas. The age of the speaker can have a large impact on the performance of the ASR system  \citep{wilpon1996study}. Having a good age classifier can help us find which age group the ASR system struggles with the most, allowing us to improve the model on that end. Additionally, the age information can provide us with clues related to age biases that the model might contain.

The age classification is a challenging task since there is no clear boundary that separates one age class from its neighbouring classes. For example, it is almost impossible to find a difference in speech between a 38-year-old person (age group 31-40) and a 41-year-old person (age group 41-50). Due to that, besides the standard accuracy metric, we also used relaxed accuracy, where the neighbouring classes are also considered as correct predictions. 

For this task, we also developed two models similarly as we did in the gender task. One operating on up to 3-second segments, and another operating on longer, up to 50-second segments. This will give as a clue about what segment lengths are sufficient for learning the task.

\begin{table}[htb]
    \centering
    \caption{Accuracy of the models on the age classification task.}
    \label{tab:age_results}
    \begin{tabular}{ccccc}
    \toprule
    \multicolumn{1}{c}{} & \multicolumn{2}{c}{\textbf{Accuracy}} & \multicolumn{2}{c}{\textbf{Relaxed Accuracy}}\\ \cmidrule(lr){2-3} \cmidrule(lr){4-5}
    \textbf{Model} & Test & Multi-transcriber test & Test & Multi-transcriber test \\
    \midrule
    3 seconds model & 33.59 & 40.16 & 79.28 & 78.48 \\
    50 seconds model & 42.39 & 52.66 & 86.34 & 89.55 \\
    \bottomrule
    \end{tabular}
\end{table}

In Table \ref{tab:age_results}, we can see the performance of both models on the test sets, using the standard and the relaxed accuracy. From the results, we can see that the model using 50 second segments performs significantly better, which indicates that more information is required for the model to learn this task. Additionally, by using the relaxed accuracy, we gained a large improvement, which suggests that most of the mistakes happen by confusing the actual class with one of the neighbouring classes.

\subsection{Dialect classification}
The participants in the Lahjoita puhetta campaign were encouraged to use their dialect and provide that information when recording the audio. Automatic dialect classification for Finnish is a challenging and underexplored task. The only previous attempt of combining audio and text modalities for Finnish dialect classification is a system combining FinBERT embeddings and a pre-trained Wav2Vec2 model, achieving good results  \citep{hamalainen2021finnish}. 
 
Since traces of dialect do not occur in every word (or even sentence), we used longer segments for the dialect classification task. We limited the samples to up to 50 seconds. The reason that we did not use the whole audio is that some samples can be multiple minutes long, which makes them hard to process.

Besides the acoustic information, for this task, we additionally experimented with enriching the input with morphological information by utilising the transcripts. To utilise both the audio and the transcript information, we used only the audio files that have corresponding transcripts. In this experiment, instead of cutting the audio to 50 seconds, we discarded the samples that are longer than that. We did so in order for the transcripts to match the audio.

To see if adding the transcripts has any benefit, we trained an audio-only model on the same samples as the model using audio and transcripts (we will refer to this as “audio subset").

Lastly, instead of using the original transcripts, we experimented with the decoded transcripts from the initial-100h TDNN model (see Section \ref{sec:hybrid}). This model is trained on the same data as the model utilising audio and transcripts, except the 100 hours used for training the ASR model. This will give us an opportunity to investigate how much the performance differs on ASR-generated transcripts and whether it is a good idea to decode the untranscribed part of the data and train the model on the whole audio and the ASR-generated transcripts.

\begin{table}[htb]
    \centering
    \caption{Accuracy of the models on the dialect classification task.}
    \label{tab:dialect_results}
    \begin{tabular}{ccc}
    \toprule
    \textbf{Model} & \textbf{Test} & \textbf{Multi-transcriber test} \\
    \midrule
    whole audio & 40.74 & 35.14 \\
    \midrule
    audio + transcripts & 32.66 & 29.20 \\
    audio + ASR transcripts & 29.19 & 30.97 \\
    audio subset & 39.73 & 38.83 \\
    \bottomrule
    \end{tabular}
\end{table}

The accuracy of the models is given in Table \ref{tab:dialect_results}. Looking at the results, we can observe that the model trained on all the audio performs better than the one trained on the audio and the available transcripts. This could indicate that the dialect information is predominant in the audio since the transcripts are not able to capture information such as pronunciation and accent. Additionally, we can observe that using the ASR transcripts degrades the performance on the test set, but it improves it slightly on the multi-transcriber test set, in comparison to using the original transcripts. This could mean that the words affected by the dialect are also difficult for the ASR model, resulting in incorrect transcriptions. Further, the audio subset model performs better than its counterpart that additionally uses the transcripts. This could indicate that instead of providing additional information, the transcripts introduce noise to the model.

Generally, the accuracy of the models is relatively low in comparison to the other metadata classification tasks. This indicates that the dialect classification in this dataset is a very difficult task and more advanced methods might be required in order to get optimal results. In general, the use of dialects in Lahjoita puhetta may be weaker and less frequent than in datasets where the particular focus on dialects may have affected the choice of participants and collection methods. 

\subsection{Topic classification}
During the collection of Lahjoita puhetta, the participants were asked to choose a theme and then talk about topics within the theme. Due to the large number of topics, we used the themes (listed in Section \ref{sec:metadata}) as labels, with the only difference being that we combined the three "Media skills" themes into one.

Similar to the dialect classification, for this task we also cut the audio segments to 50 seconds and trained an audio-only model on the whole data.

Topic classification is often done on text. For that purpose, we trained a text-only model on the samples that are 50 seconds or less. Additionally, we tried utilising the acoustic and the morphological information by processing the audio and the transcripts together, just like in the dialect classification task.
Furthermore, we investigated the performance of the model, when provided with ASR decoded transcripts, instead of the gold-standard ones. The ASR transcripts are generated using the same initial-100h TDNN model as the one in the dialect classification task.

Lastly, we investigated the effect of the audio on the topic classification task. For that purpose, we developed an audio-only model that is trained on the same data as the models using the original transcripts.

\begin{table}[htb]
    \centering
    \caption{Accuracy of the models on the topic classification task.}
    \label{tab:topic_results}
    \begin{tabular}{ccc}
    \toprule
    \textbf{Model} & \textbf{Test} & \textbf{Multi-transcriber test} \\
    \midrule
    whole audio & 65.65 & 73.93 \\
    \midrule
    transcripts & 82.06 & 88.65 \\
    ASR transcripts & 82.06 &  86.52 \\
    audio + transcripts & 81.34 & 87.94 \\
    audio + ASR transcripts & 80.14 & 86.52 \\
    audio subset & 57.10 & 69.47\\
    \bottomrule
    \end{tabular}
\end{table}

The results for the topic classification task are given in Table \ref{tab:topic_results}. From the table, we can observe that the model that uses the original transcripts achieves slightly better results than the one using the ASR-generated transcripts on the multi-transcriber test set, whereas on the test set, they perform identically. Additionally, the models using only the transcripts achieve significantly better results than the model using the whole audio, even though the audio-only model was trained on far more data. When jointly using the audio and the transcript information, we can see that there is a small degradation in comparison to using only the transcripts. This could indicate that the audio does not provide any additional information that would help the model. Another thing to consider is that the audio encoder that we are using is quite small, so a bigger model might be necessary if we want to benefit more from the acoustic information. When we combined the audio and the ASR-generated transcripts, we observed only a small degradation in the performance, in comparison to using the audio with the original transcripts. This could indicate that certain keywords affect the topic classification and the ASR system is good at detecting them. Using this knowledge, in future experiments we can generate transcripts for the untranscribed part of the data and use them in addition to the audio, to train a big model that utilises audio and transcript information.  From the results obtained on the model trained on the subset of the audio, we can see that there is a significant degradation in the results in comparison to the model that uses only the transcripts. This confirms that the textual information content is sufficiently dense for this task. Generally, the models were able to learn the task relatively well, while still leaving some space for improvement, especially on the audio side.

\subsection{Analysis of metadata classification errors}
\begin{figure}[htb]
\includegraphics[width=\textwidth]{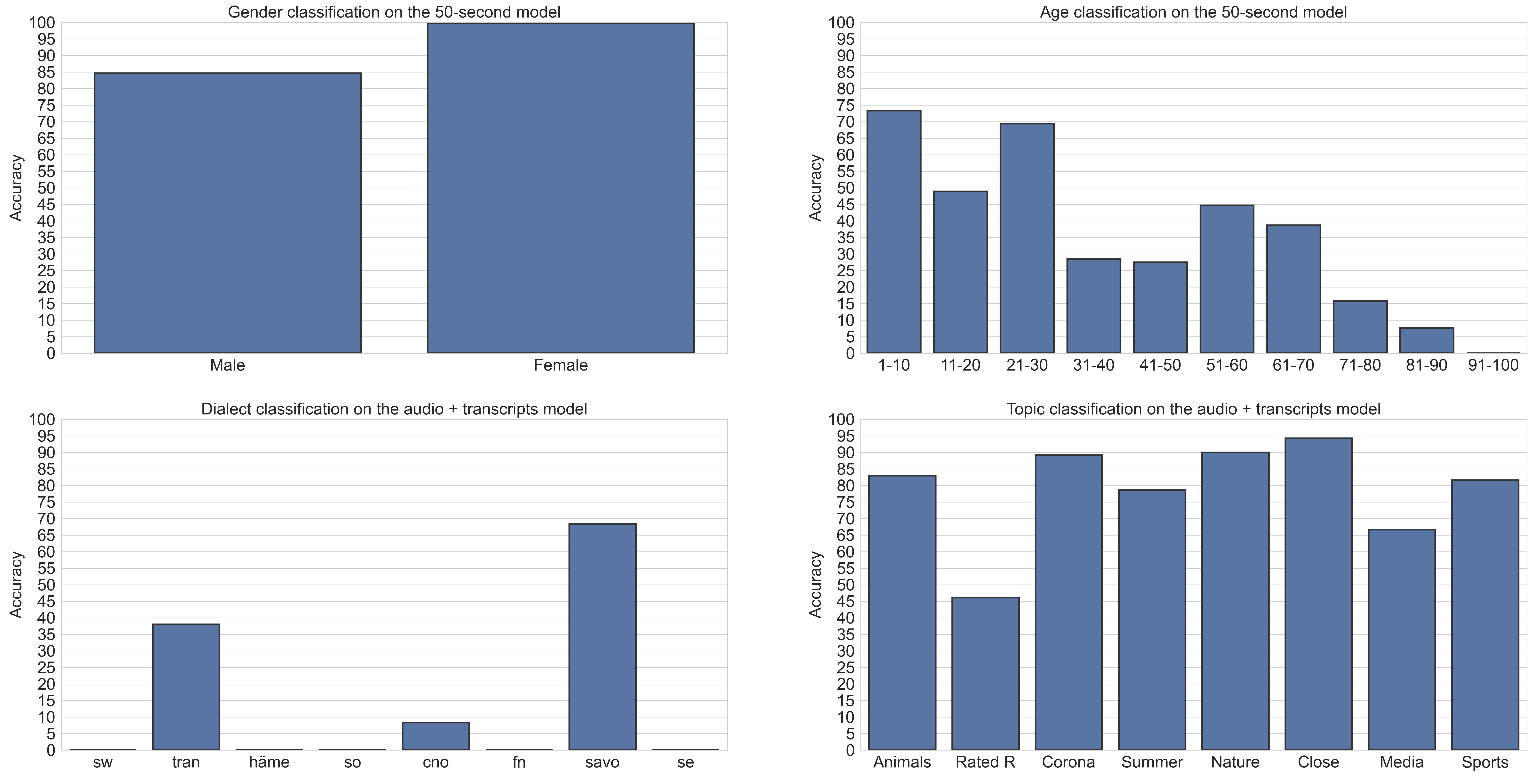}
\caption{Metadata accuracy per class on the test set.}
\label{fig:metadata_acc_by_class}
\end{figure}

\begin{figure}[htb]
\includegraphics[width=\textwidth]{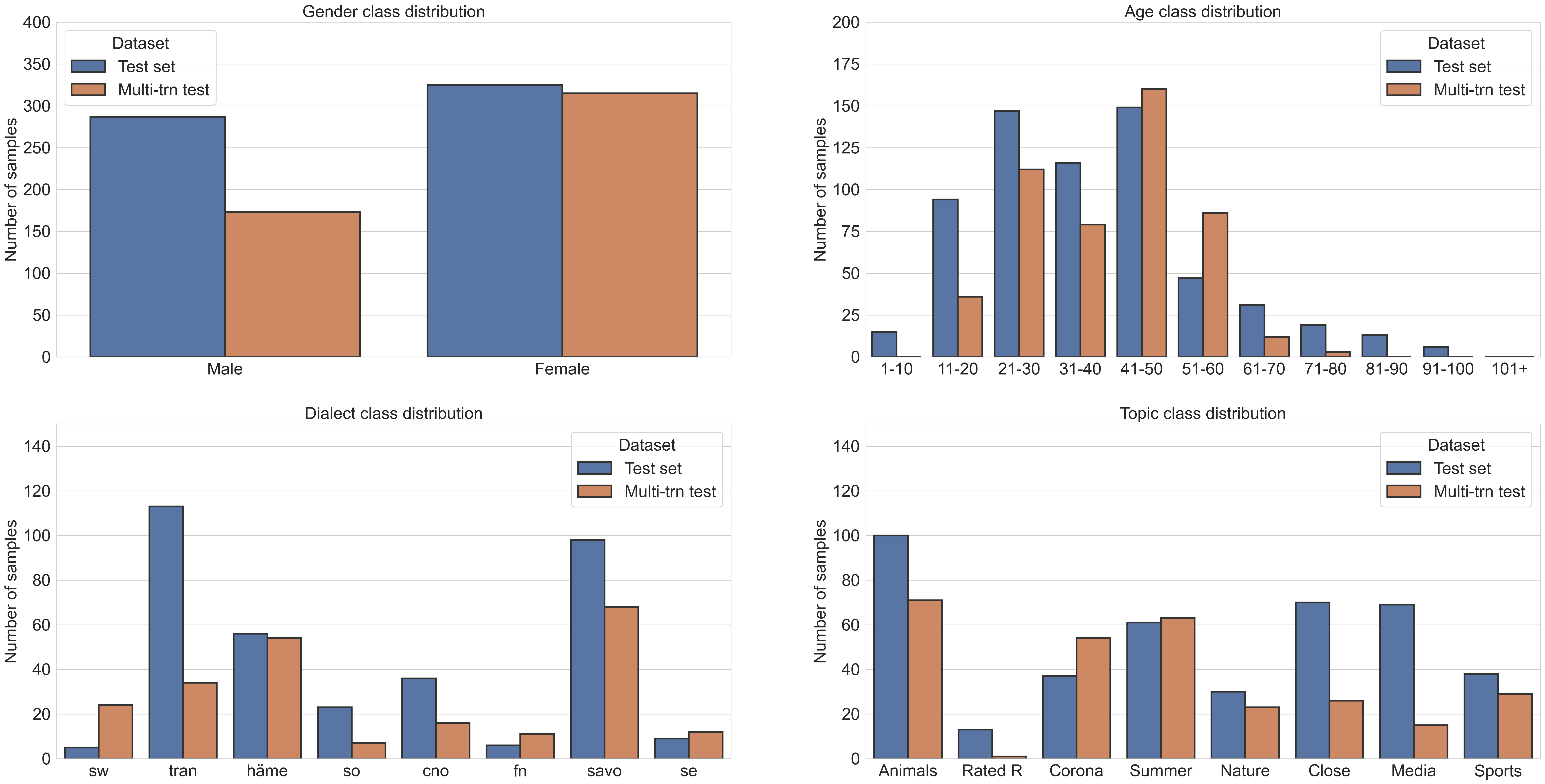}
\caption{Metadata class distribution for the test sets.}
\label{fig:metadata_class_dist}
\end{figure}

To further investigate which classes are challenging for the metadata classification models, we evaluated them on each class individually. The results of the analysis are given in Figure \ref{fig:metadata_acc_by_class}. Additionally, in Figure \ref{fig:metadata_class_dist} we can observe the number of samples per class that were used during the evaluation.

In the gender classification case, we can see that the model performs significantly better on the female examples. The reason could be that there is a high disproportion between male and female samples in the training set.

On the age classification task, we can observe that the model performs better on the lower age groups and struggles with the elderly, especially the ones in the 91-100 age group, where the model misclassified all the samples.

\begin{figure}[htb]
    \centering
    \includegraphics[height=7cm, width=9cm]{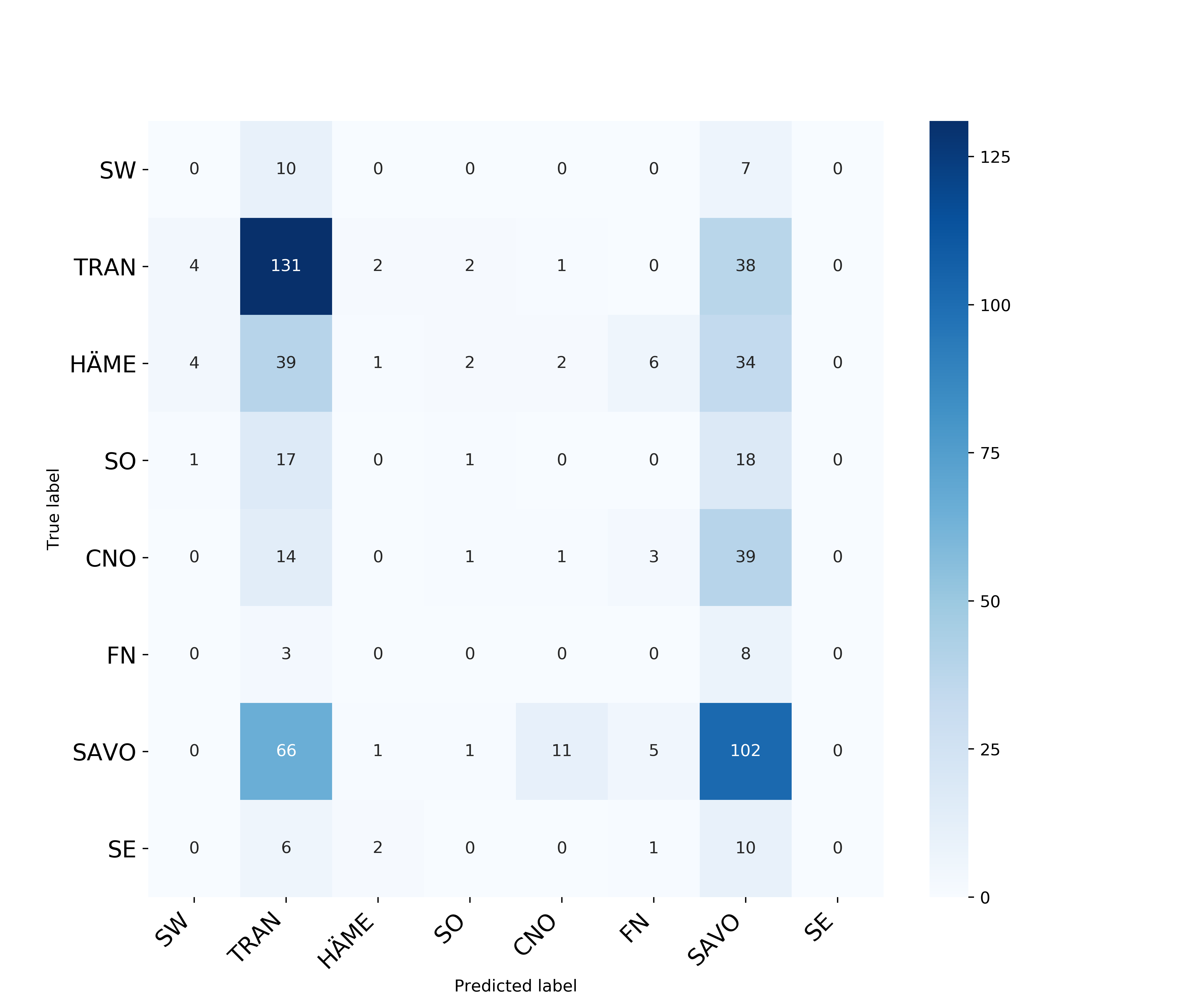} 
    \caption{Confusion matrix for the dialect classification model on the test set.}
    \label{fig:dialect_confusion}
\end{figure}

On the dialect classification plot, we can see that the model misclassified all the samples from several dialect groups. This is not surprising, considering that many of those dialect groups have only a couple of samples and the general accuracy of the model is low. To further investigate the mistakes that the model made on this task, we plotted a confusion matrix, presented in Figure \ref{fig:dialect_confusion}. From the matrix, we can see that the HÄME dialects are mostly confused with TRAN and SAVO, which are neighbouring dialects in our dialect grouping. Similar observations can be made with the CNO dialect group, which is mostly confused with its neighbouring SAVO group.

On the topic classification plot, we can see that the model is performing well on almost all the classes. The weakest one seems to be the Rated R class, which generally has a low number of samples in the training and testing sets.

\section{Possible future directions}

Although we have demonstrated with multiple use-cases the usefulness of the collected corpus in this article, there are still numerous possibilities to utilise the dataset. Out of those possibilities, we plan to realise a few in the near future. Perhaps the most evident utilisation of the corpus is a speaker recognition system. The large number of speakers of various ages speaking different dialects would enable us to build a robust and accurate model for Finnish data. 
The carefully transcribed portion of the data would make an interesting resource for colloquial Finnish text to speech (TTS) systems. We hypothesise that the marked non-speech parts and disfluencies could be leveraged to create a more natural TTS that can hesitate, restart words, and make non-speech sounds at the appropriate places. The AED ASR experiments uncover clear difficulties with the long-form recordings in this dataset. The results on an ad-hoc segmented version of the development data portion were on par with the HMM/DNN systems' results on the official data, which suggests that solving these technical difficulties would make AED systems viable approach for this data.
The last future direction that we wish to mention concern the untranscribed part of the dataset. We have already demonstrated that it can be used for semi-supervised learning, and we plan to investigate its usefulness with other self-supervised and unsupervised methods. Specifically, we intend to build a truly Finnish Wav2Vec2 model, which would be pre-trained on purely Finnish data and fine-tuned with the large transcribed part of the corpus.

A similar effort for large-scale collection of donated speech for other languages such as the second national language of Finland, i.e. the variety of Swedish spoken in Finland is already on-going. Efforts for applying this collection concept and tools for collecting minority languages spoken in Finland are also planned.

\section{Conclusions}

In this paper, we presented a new, large-scale, conversational Finnish speech corpus. The 3600 hours, out of which 1600 are transcribed, include over twenty thousand speakers from all age groups and from all the regions of Finland. To ensure the high quality of the transcripts, the transcribers were thoroughly evaluated using manual and automatic techniques. The techniques for data collection and annotation applied in this paper provide a valuable resource for future similar attempts at collecting large-scale data.

To establish that the \textit{Lahjoita puhetta} data is useful for training ASR systems, we built several hybrid HMM/DNN and end-to-end baseline models and made them publicly available. The varied ASR experiments, with the best system achieving 21.75\% WER, showed that the data is suitable for building such systems. This should also be compared with the initial word-level interannotator disagreement ranging between 13-20\% on this data type. Furthermore, the large untranscribed part of the corpus can be utilised for unsupervised and semi-supervised training. 
The rich metadata provided by the participants allowed us to successfully train various metadata classification models, demonstrating further use-cases for the dataset. The benchmark metadata classification models are publicly released together with the ASR models.

The large and diverse \textit{Lahjoita puhetta} dataset will be freely available for research purposes, and for commercial use at a low cost. We hope this encourages researchers and companies to further develop  language technologies and bridge the gap between research and commercial use.

\section*{Acknowledgements}
We are grateful for the Academy of Finland project funding number 337073 “FIN-CLARIN as a Collaborative Platform for Speech Processing” and the Academy of Finland grant 329267 in Digital Humanities programme's project “Movie Making Finland: Finnish fiction films as audiovisual big data, 1907–2017”. The computational resources were provided by Aalto ScienceIT.

\section*{Statements and Declarations}
The authors have no relevant financial or non-financial interests to disclose.

%
%


\bibliographystyle{spbasic}      
\bibliography{references}   

%
%

\appendix
\section{Transcriber instructions} \label{ap:transcriber_instructions}
The transcribers were given the following guidelines: 

Speech produced continuously is written on the same line without line breaks. A line break is marked at each point where the speaker clearly pauses. In addition to line breaks, breaks or their durations are not marked or separated in any other way. The 
Finnish alphabet (a-zåäö) is used to record the verbal content of speech. Normal spelling punctuation, such as periods, commas, or question marks, is not included in the transcription. Numbers are not used either. Words are written as accurately as possible in the exact form in which the speaker produced them, e.g., write “pitsaa”, not “pizzaa”. Do not attempt to correct any mistakes made by the speaker. Do not add any additional comments to the text. All words can be written in lower case. If the word is clearly a proper noun, a capital letter may be used. However, words beginning an utterance or a sentence are not capitalised.

Hyphens or periods are used for punctuation only in the following special cases: A hyphen indicates, for example, a missed or “incorrectly started” word, e.g. “predi- president” or a word from which only the remainder can be heard, or from which the speaker speaks only the remainder: “-sident.” In the case of a compound whose suffix ends in the same vowel in which the suffix begins, a hyphen may be used between the parts: “tila-autolla” ("with a minivan").

When there is a point in the speech where the speaker makes vague fill-in or hesitation sounds, pouts, coughs, laughs, yawns, or sighs so that the sound is clearly heard, and the speaker does not produce the speech at the same time, the sound can be marked with, for example: \textit{.fp} (filled pause) can mark a complex or ambiguous fill or hesitation sound that is not sufficient to describe “mm”, “aa” or “öö”, \textit{.ct} (clear throat), \textit{.cough} (coughing), \textit{.laugh} (laughing), \textit{.yawn} (yawning), \textit{.sigh} (sigh, loud inhalation and exhalation), \textit{.br} (breath, single clearly audible in- or exhalation sound).

However, if the speaker, for example, laughs or yawns while speaking, do not try to include the laughter or yawn in the transliteration of the speech (for example, using the letters h). In such situations, precisely transcribing is not useful for the purpose of the material. The aim is to transcribe only the verbal content of the speech and, if necessary, the sounds to be heard between the words.


\end{document}